\documentclass[conference]{IEEEtran}
\IEEEoverridecommandlockouts
\usepackage{cite}
\usepackage{amsmath,amssymb,amsfonts}
\usepackage{algorithmic}
\usepackage{graphicx}
\graphicspath{{figures/}}
\usepackage{textcomp}
\usepackage{xcolor}
\usepackage{parskip}

\def\BibTeX{{\rm B\kern-.05em{\sc i\kern-.025em b}\kern-.08em
    T\kern-.1667em\lower.7ex\hbox{E}\kern-.125emX}}
\begin{document}

\title{From implicit learning to explicit representations\\
}

\makeatletter
\newcommand{\linebreakand}{%
  \end{@IEEEauthorhalign}
  \hfill\mbox{}\par
  \mbox{}\hfill\begin{@IEEEauthorhalign}
}
\makeatother

\author{
\IEEEauthorblockN{Naomi Chaix-Eichel}
\IEEEauthorblockA{
\textit{Inria Bordeaux Sud-Ouest}\\
\textit{Univ. Bordeaux, CNRS } \\
%\textit{naomi.chaix-eichel@inria.fr}
}
\and
\IEEEauthorblockN{Snigdha Dagar}
\IEEEauthorblockA{
\textit{Inria Bordeaux Sud-Ouest}\\
\textit{Univ. Bordeaux, CNRS } \\
%\textit{snigdha.dagar@inria.fr}
}
\and
\IEEEauthorblockN{Quentin Lanneau}
\IEEEauthorblockA{
\textit{Inria Bordeaux Sud-Ouest}\\
\textit{Univ. Bordeaux, CNRS } \\
}
\and
\IEEEauthorblockN{Karen Sobriel}
\IEEEauthorblockA{
\textit{Inria Bordeaux Sud-Ouest}\\
\textit{Univ. Bordeaux, CNRS} \\
}
\linebreakand
\IEEEauthorblockN{Thomas Boraud}
\IEEEauthorblockA{
\textit{Univ. Bordeaux, CNRS } \\
%\textit{thomas-boraud@u-bordeaux.fr}
}

\and
\IEEEauthorblockN{Frédéric Alexandre}
\IEEEauthorblockA{
\textit{Inria Bordeaux Sud-Ouest}\\
\textit{Univ. Bordeaux, CNRS } \\
%\textit{nicolas.rougier@inria.fr}
}
\and
\IEEEauthorblockN{Nicolas Rougier}
\IEEEauthorblockA{
\textit{Inria Bordeaux Sud-Ouest}\\
\textit{Univ. Bordeaux, CNRS } \\
%\textit{nicolas.rougier@inria.fr}
}
}

\maketitle

\begin{abstract}
Using the reservoir computing framework, we demonstrate how a simple model can solve an alternation task without an explicit working memory. To do so, a simple bot equipped with sensors navigates inside a 8-shaped maze and turns alternatively right and left at the same intersection in the maze. The analysis of the model's internal activity reveals that the memory is actually encoded inside the dynamics of the network. However, such dynamic working memory is not accessible such as to bias the behavior into one of the two attractors (left and right). To do so, external cues are fed to the bot such that it can follow arbitrary sequences, instructed by the cue. This model highlights the idea that procedural learning and its internal representation can be dissociated. If the former allows to produce behavior, it is not sufficient to allow for an explicit and fine-grained manipulation.
\end{abstract}


\section{Introduction}

Suppose you want to study how an animal, when presented with two options A and B, can learn to alternatively choose A then B then A, etc. One typical lab setup to study such alternate decision task is the T-maze environment where the animal is confronted to a left or right turn and can be subsequently trained to display an alternate choice behavior. This task can be easily formalized using a block world as it is regularly done in the computational literature. Using such formalization, a simple solution is to negate (logically) a one bit memory each time the model reaches A or B such that, when located at the choice point, the model has only to read the value of this memory in order to decide to go to A or B. 
However, as simple as it is, this abstract formalization entails the elaboration of an explicit internal representation keeping track of the recent behavior, implemented in a working memory that can be updated when needed.
% I proposed a simpler version, instead of the previous one: 
%However, as simple as it is, this abstract formalization entails an explicit representation of the two options A and B (that we assume the animal is also using) as well as the need for an explicit working memory that can be updated when needed. 
But then, what could be the alternative? Let us consider a slightly different setup where the T-Maze is transformed into a closed 8-Maze (see figure \ref{fig:t-maze}-Left). Suppose that you can only observe the white area when the animal is evolving along the arrowed line (both in observable and non-observable areas). From the observer point of view, the animal is turning left one time out of two and turning right one time out of two. Said differently, the observer can infer an alternating behavior because of its partial view of the system. The question is: does the animal really implement an explicit alternate behavior or is it merely following a mildly complex dynamic path? This is not a rhetorical question because depending on your hypothesis, you may search for neural correlates that actually do not exist. Furthermore, if the animal is following such mildly complex dynamic path, does this mean that it has no explicit access to (not to say no consciousness of) its own alternating behavior?
\begin{figure}
    \centering
    \includegraphics[width=\columnwidth]{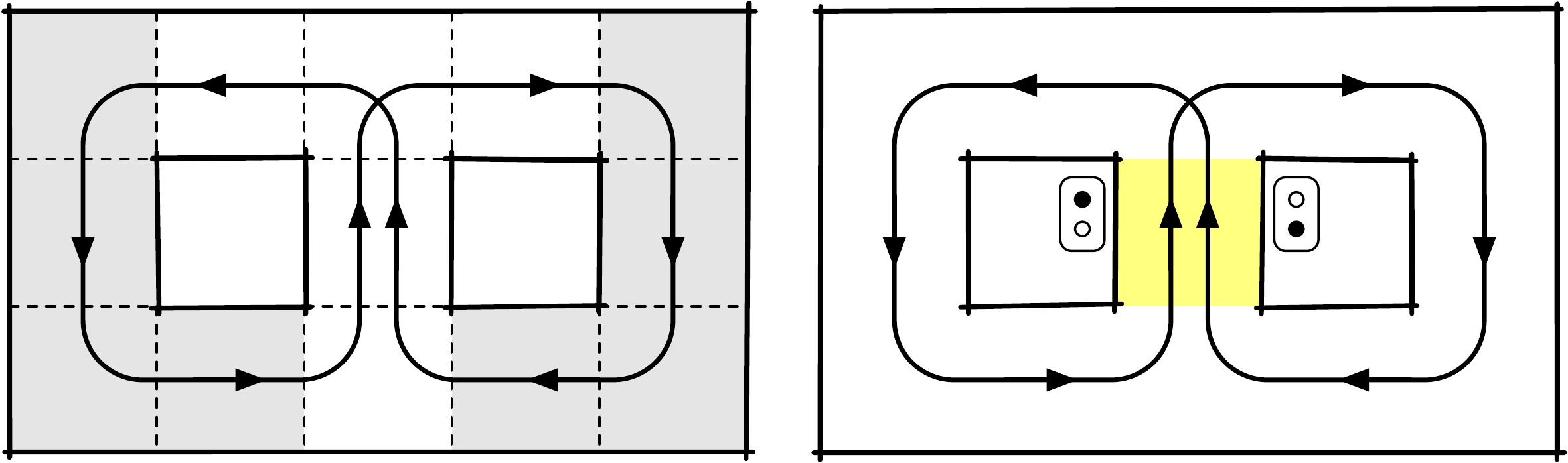}
    \caption{Left: An expanded view of a T-Maze. An observer can infer an alternating behavior because of her partial view (white area) of the system. Right: 8-maze with cues. A cue (left or right) is given only when the bot is present inside the yellow area.}
    \label{fig:t-maze}
\end{figure}

This question is tightly linked to the distinction between implicit learning (generally presented as sub-symbolic, associative and statistics-based) and explicit learning (symbolic, declarative and rule-based). Implicit learning refers to the {\em non-conscious effects that prior information processing may exert on subsequent behavior} \cite{Cleeremans:2009}. It is implemented in associative sensorimotor procedural learning and also in model-free reinforcement learning, with biological counterparts in the motor and premotor cortex and in the basal ganglia. Explicit learning is associated with consciousness or awareness, and to the idea of building explicit mental representations \cite{cleeremans2007consciousness} that can be used for flexible behavior, involving the prefrontal cortex and the hippocampus. This is what is proposed in model-based reinforcement learning and in other symbolic approaches for planning and reasoning. These strategies of learning are not independent but their relations and interdependencies are not clear today. Explicit learning is often observed in the early stages of learning whereas implicit learning appears on the long run, which can be explained as a way to decrease the cognitive load. But there is also a body of evidence, for example in sequence learning \cite{clegg1998sequence} or artificial grammar learning studies \cite{Reber:1996}, that suggests that explicit learning is not a mandatory early step and that improvement in task performance are not necessarily accompanied by the ability to express the acquired knowledge in an explicit way \cite{dienes1997implicit}.  

Coming back to the task mentioned above, it is consequently not clear if we can learn rules without awareness and then to what extent can such implicit learning be projected to performance in an unconscious way? Furthermore, without turning these implicit rules into an explicit mental representation, 
is it possible to manipulate the rules, which is a fundamental trademark of flexible adaptable control of behavior? 

%There are a number of computational models that describe the neurobiology behind cognitive control (this flexible explicit control of behavior evoked above) by using symbolic representations of the task. The question of \textit{how} behaviour is made available to conscious, cognitive control remains unresolved on both computational and neuroscience levels. Traditional reinforcement learning (RL) methods are insufficient to learn rules which do not possess the Markovian property: they need explicit internal representations to serve as a working memory that can store previous experience. 

Using the reservoir computing framework generally considered as a way to implement implicit learning, we first propose that a simple alternation or sequence learning task can be solved without an explicit pre-encoded representation of memory. However, to then be able to generate a new sequence or manipulate the rule learnt, we explain that inserting explicit cues in the decision process is needed. In a second series of experiments, we provide a proof of concept still using the reservoir computing framework, for the hypothesis that the recurrent network forms contextual representations from implicitly acquired rules over time. We then show that these representations can be considered explicit and necessary to be able to manipulate behaviour in a flexible manner.

In order to provide preliminary interpretation of what is observed here, it is reminded that recurrent networks, particularly models using the reservoir computing framework, are a suitable candidate to model the prefrontal cortex\cite{dominey1995model}, also characterized by local and recurrent connections. Given their inherent sensitivity to temporal structure, it also makes these networks adaptable for sequence learning. This approach has been used to model complex sensorimotor couplings \cite{tani1999learning} from the egocentric view of an agent (or animal) that is situated in its environment and can autonomously demonstrate reactive behaviour from its sensory space\cite{Antonelo:2015}, as we also do in the first series of experiments, for  learning sensorimotor couplings by demonstration, or imitation. In the second series of experiment, we propose that the prefrontal cortex is the place where explicit representations can be elaborated when flexible behaviors are required.

\section{Methods and Task}

The objective is the creation of a reservoir computing network of type Echo State Network (ESN) that controls the movement of a robot \cite{Antonelo:2012,Antonelo:2015}, which has to solve a decision-making task (alternately going right and left at an intersection) in the maze presented in figure \ref{fig:t-maze}.

%As \cite{cleeremans2007consciousness} presented, we want to introduce the notion of meta-representation that goes with the emergence of ”consciousness”: first, once the ESN trained, the agent is able to do the task. It corresponds to what we call the "first-order representation". Second, if the agent is "conscious" of this ability, i.e. it can "observe", access and manipulate this internal representation, it corresponds to the second-order representation. Our purpose is to reproduce this two-level representation, with the first one corresponding to the ESN that resolves the task in the maze, the second one corresponding to another network that takes as input the states of the ESN and predict its output or solve a secondary task derived from the initial one.  In this paper, we will introduce the first-order network and its task. We show that after training, the model is not only capable of reproducing what it learnt, but can also achieve more complex tasks if we feed its input with proper contextual orders to execute.

\subsection{Model Architecture : Echo State Network}

An ESN is a recurrent neural network (called reservoir) with randomly connected units, associated with an input layer and an output layer, in which only the output (also called readout) neurons are trained. The neurons have the following dynamics:
\begin{eqnarray}
{\bf x}[n] &=& (1-\alpha){\bf x}[n-1] + \alpha \tilde{{\bf x}}[n]\\
\tilde{\bf x}[n] &=& \tanh(W {\bf x}[n-1] + W_{in}[1;{\bf u}[n]])\\
{\bf y}[n] &=& f(W_{out}[1;\tilde{\bf x}(n)])
\end{eqnarray}
where \({\bf x}(n)\) is a vector of neurons activation, \(\tilde{\bf x}(n)\) its update, \({\bf u}(n)\) and \({\bf y}(n)\) are respectively the input and the output vectors, all at time \(n\).  \(W\),  \(W_{in}\),  \(W_{out}\) are respectively the reservoir, the input and the output weight matrices. The notation \([.;.]\) stands for the concatenation of two vectors. \(\alpha\) corresponds to the leak rate. 
\(tanh\) corresponds to the hyperbolic tangent function and \(f\) to linear or piece-wise linear function. \\

The values in \(W\),  \(W_{in}\),  \(W_{out}\) are initially randomly chosen. While \(W\),  \(W_{in}\) are kept fixed, the output weights \(W_{out}\) are the only ones plastic (red arrows in Figure \ref{fig:model1}). In this model, the output weights are learnt with the ridge regression method (also known as Tikhonov regularization):
\begin{equation}
W_{out} = Y^{target}X^T(XX^T + \beta I)^{-1}
\end{equation}
where \(Y^{target}\) is the target signal to approximate, \(X\) is the concatenation of 1, the input and the neurons activation vectors: \([1;u(n);x(n)]\), \(\beta \) corresponds to the regularization coefficient and \(I\) the identity matrix. \\

\subsection{Experiment 1 : Uncued sequence learning}

The class of tasks called spatial alternation has been widely used to study hippocampal and working memory functions \cite{frank}. For the purpose of our investigation, we simulated a continuous version of the same task, wherein the agent needs to alternate its choice at a decision point, and after the decision, it is led back to the central corridor, in essence following an 8-shaped trace while moving (see figure \ref{fig:t-maze}-Left). This alternation task is widely believed to require a working memory such as to remember what was the previous choice in order to alternate it. Here we show that the ESN previously described is sufficient to learn the task without an explicit representation of the memory. 

\subsubsection{Tutor model}
In order to generate data for learning, we implemented a simple Braintenberg vehicle where the robot moves automatically with a constant speed and changes its orientation according to the values of its sensors. At each time step the sensors measure the distance to the walls and the bot turns such as to avoid the walls. At each timestep, the position of the bot is updated as follows:
\begin{eqnarray}
\theta(n) &=& \theta(n-1) + 0.01 \sum_i \alpha_is_i\\
p(n) &=& p(n-1) + 2*(cos(\theta(n)) + sin(\theta(n)))
\end{eqnarray}
where \(p(n)\) and \(p(n+1)\) are the positions of the robot at time step \( n\)  and \(n+1\) , \( \theta(n)\) is the orientation of the robot, calculated as the weighted sum ($\alpha_i$) of the values of the sensors~$s_i$. The norm of the movement is kept constant and fixed at~2. 
% Based on the Braitenberg algorithm, the ESN predicts the next orientation of the robot according to the values of the sensors received as input (see figure \ref{fig:model1}).
%
\begin{figure}
    \centering
    \includegraphics[width=\linewidth]{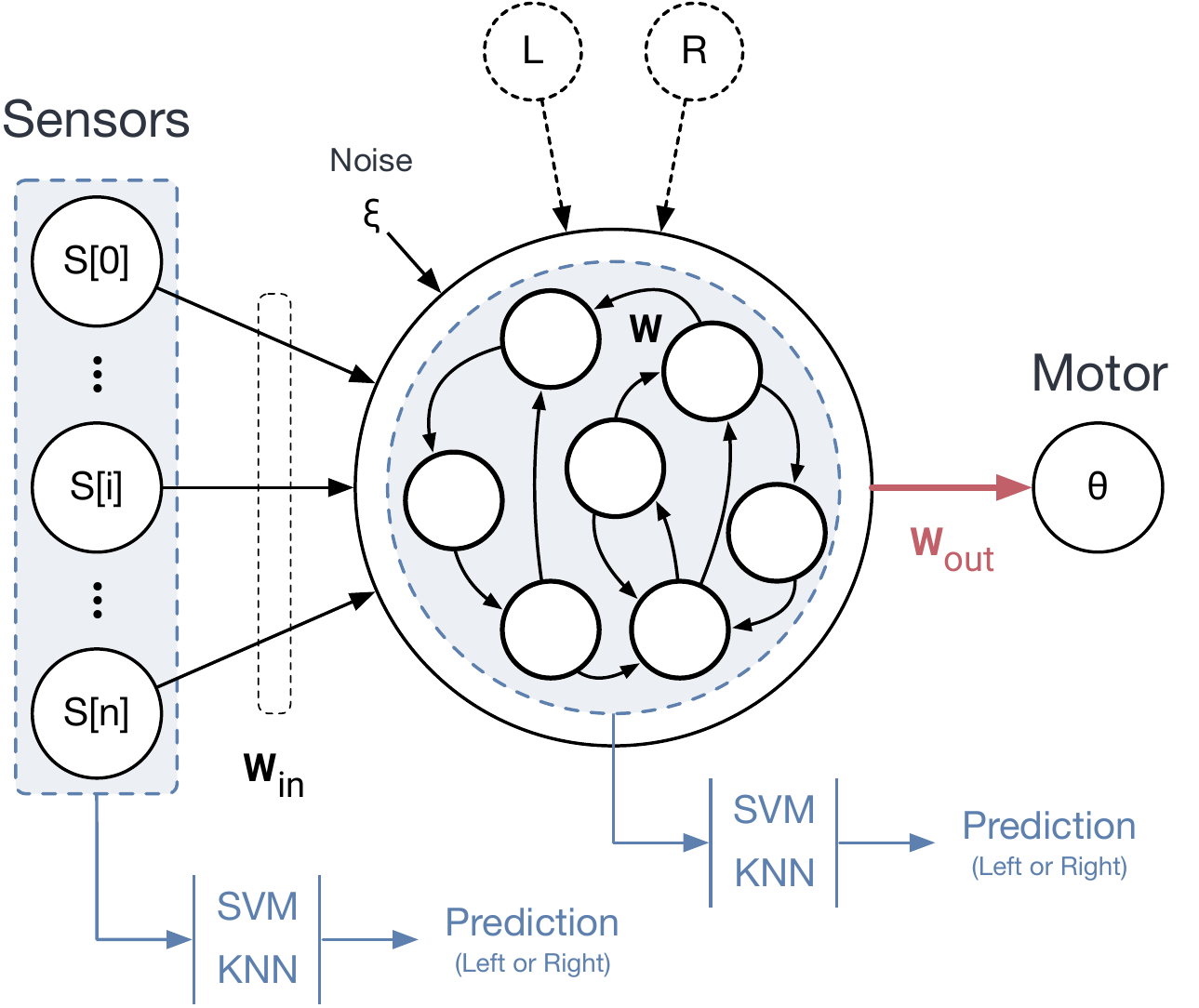}
    \caption{Model Architecture with 8 sensor inputs, and a motor output (orientation). The black arrows are fixed while the red arrows are plastic and are trained. The reservoir states are used as the input to a classifier which is trained to make a prediction about the decision (going left or right) of the robot. A left (L) and right (R) cue can be fed to the model depending on the experiment (see Methods). }
    \label{fig:model1}
\end{figure}

% @ Nicolas : can we add the context inputs also to this schematics ? I removed the other figure

\subsubsection{Training data}

The ESN is trained using supervised learning, containing samples from the desired 8-shaped trajectory. 
Since the Braitenberg algorithm only aims to avoid obstacles, the robot is forced into the desired trajectory by adding walls at the intersection points as shown in figure \ref{fig:invis_w}. After generating the right pathway, the added walls are removed and the true sensor values are gathered as input. Gaussian noise is added to the position values of the robot at every time step in order to make the training more robust. Approximately 50,000 time steps were generated (equivalent to 71 complete 8-loops) and separated into training and testing sets.
\begin{figure}
    \centering
    \includegraphics[width=\linewidth]{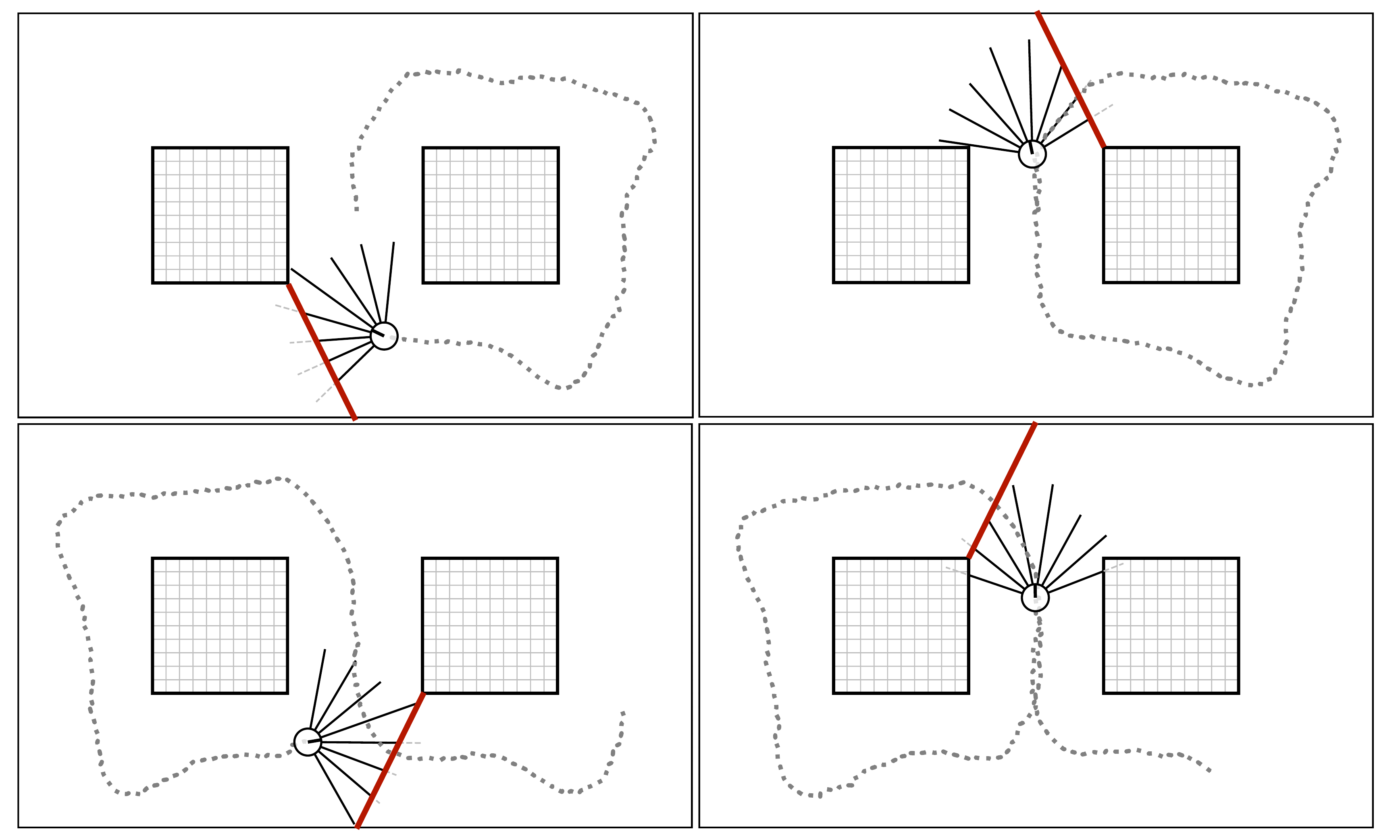}
    \caption{Generation of the 8-shape pathway with the addition of walls at the intersection points}
    \label{fig:invis_w}
\end{figure}

\subsubsection{Hyper parameters tuning}

The ESN was built with the python library ReservoirPy \cite{Trouvain2020} with the hyper-parameters presented in table \ref{tab:params}, column "Without context". The order of magnitude of the hyper-parameters was first found using the Hyperopt python library \cite{bergstra2015hyperopt}, then these were fine tuned manually. The ESN receives as input the values of the 8 sensors and output the next orientation. 
\begin{table}
    \centering
    \begin{tabular}{l l l}
    \hline
    Parameter & Without context & With context \\
    \hline
    Input size &  8 & 10\\
    Output size & 1 & 1\\
    Number of units & 1400 & 1400\\
    Input connectivity  & 0.2  & 0.2\\
    Reservoir connectivity & 0.19 & 0.19\\
    Reservoir noise & 1e-2 & 1e-2 \\
    Input scaling & 1 & 1(sensors), 10.4695 (cues)\\
    Spectral Radius & 1.4 & 1.505\\
    Leak Rate & 0.0181 & 0.06455\\
    Regularization & 4.1e-8 & 1e-3 \\
    \hline
    \end{tabular}
    \caption{Parameter configuration for the ESN}
    \label{tab:params}
\end{table}

\subsubsection{Model evaluation}

The performance of the ESN has been calculated with the Normalized Root Mean Squared Error metrics (\(NRMSE\)) and the R square (\(R^2\)) metrics, defined as follows :

\begin{equation}
NRMSE = \frac{\sqrt{\frac{\sum_{i=1}^{n} (y_i - \hat{y_i})^2}{n}}}{\sigma}
\label{eq:nrmse}
\end{equation}

\begin{equation}
R^2 = 1- \frac{\sum(y_i-\hat{y_i})^2}{\sum(y_i-\bar{y})^2}
\label{eq:r2}
\end{equation}

where \(y_i\), \(\hat{y_i}\) and \(\bar{y}\) are respectively the desired output, the predicted output and the mean of the desired output.

\subsubsection{Reservoir state analysis}

In this section the reservoir states are analysed such as to inspect to which extent they form an internal and hidden representation of the memory. To do so, we use Principal Component Analysis (PCA), a dimensionality reduction method enabling the identification of patterns and important features of the processed data. PCA is carried out on the reservoir states for each position of the robot during the 8-shape trajectory.
We continued the analysis by doing a classification of the reservoir states. We made the assumption that it is possible to know the future direction of the robot observing the internal states of the reservoir. This implies that the reservoir states can be classified in two classes: one related to the prediction of going left and the other related to the prediction of going right. Two standard classifiers, the KNN (K-Nearest Neighbors) and the SVM (Support Vector Machine) were used. They take as input the reservoir state at each position of the bot while executing the 8-shape and predict the decision the robot will take at the next intersection (see figure \ref{fig:model1}). Since the classifiers are trained using supervised learning, the training data were generated in the central corridor of the maze (yellow area in figure \ref{fig:t-maze}-Right), assuming that it is where the reservoir is in the state configuration in which it already knows which direction it will take at the next intersection. 900 data points were generated and separated into training and testing sets. 

\subsection{Experiment 2 : 8 Maze Task with Contextual Inputs}
% In the first experiment, the robot controlled by the reservoir has acquired the right implicit rules over time so as to reproduce the 8-shape trajectory. The reservoir solved the simple alternation sequence thanks to its internal representation of the memory and without using any explicit representation of the task. However, due to its lack of internal memory capacity, the reservoir is not able to generate new and more complex sequences. The alternative is then to send cues to the robot while it is moving in the maze to explicitly associate its internal state to its behavior, which will force it to take a specific direction and make more complex sequences. 
In this experiment, we fed the reservoir with two additional inputs that represent the next decision, one being related to a right turn (R) and the other to a left turn (L) (see figure \ref{fig:model1}).
%
%these cues correspond to two additional inputs of the reservoir (one related to right, the other related to left), and are called here "contextual inputs" (see the model architecture in figure \ref{fig:model1}).
%
They are binary values, switched to a value of 1 only when the bot is known to take the corresponding direction. We thus built a second ESN with the hyper-parameters presented in TABLE \ref{tab:params}, column "With context". The network is similar to the previous one, except that the contextual inputs are added with a different input scaling than the one used for the sensors inputs. During the data generation, the two additional inputs are set to 0 everywhere in the maze, except in the central corridor.

\section{Results}

\subsection{Motor sequence learning}
We first show that a recurrent neural network like the ESN can learn a rule-based trajectory in the continuous space, without an explicit memory or feedback connections. The score of the ESN is shown in TABLE \ref{tab:score1} and the results for the trajectory predicted by the ESN are presented in figure \ref{fig:8traj} and in the top panel in figure \ref{fig:positions}. At each period of about 350 steps, a behavior or decision switch takes place, which is evident from the crests and troughs in the y-axis coordinates. It can be seen that the ESN correctly predicts the repeated alternating choice in the central arm of the maze. In addition to switching between the left and right loops, the robot also moves through the environment without colliding into obstacles. 

\begin{table}[!h]
    \centering
    \begin{tabular}{l l l l}
    \hline
    \multicolumn{4}{l}{Performance of the ESN for 50 simulations}\\
    \hline
    \multicolumn{2}{l}{\(NRMSE\)} & \multicolumn{2}{l}{\(R^2\)} \\
    \hline
    Mean &  0.0171   &  Mean &   0.9962 \\
    Variance & 5.4466e-06  & Variance & 1.0192e-06 \\
    \hline
    \end{tabular}
    \caption{\(NRMSE\) and \(R^2\) score of the ESN with 8 inputs }
    \label{tab:score1}
\end{table}

\begin{figure}
    \centering
    \includegraphics[width=\linewidth]{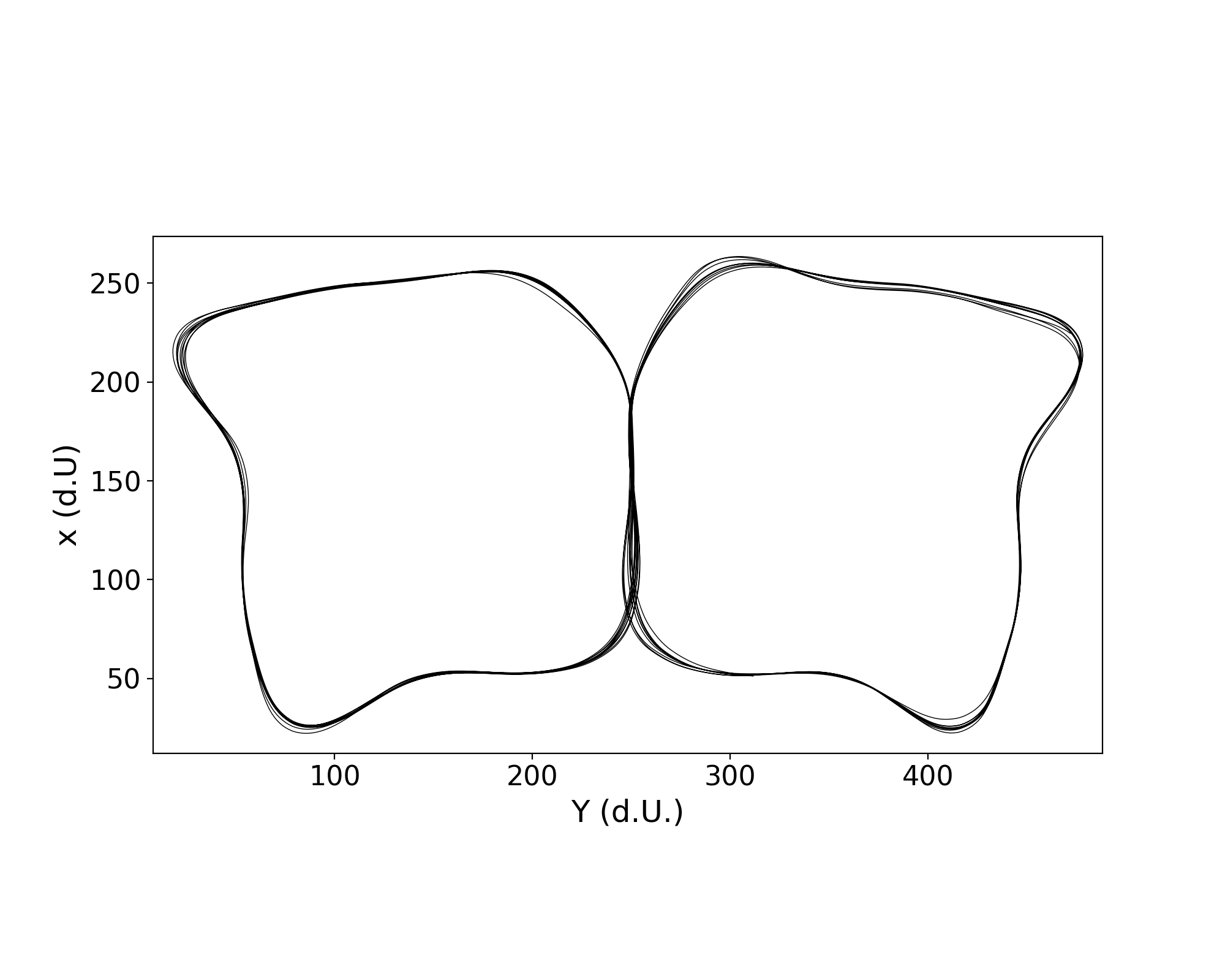}
    \caption{The trajectory of the robot following the 8-trace in the cartesian map.}
    \label{fig:8traj}
\end{figure}

\subsection{Reservoir State Prediction}

Next, we show that even a simple classifier such as SVM or KNN can observe the internal states of the reservoir and learn to predict the decision (whether to go left or right) of the network. The results of the predictions are presented in the top part of figure \ref{fig:pred_sensors}. As expected, there is a periodicity of choice in line with the position of the bot in the maze, showing that the classification is relevant. At each time step, both classifiers output the same prediction with a small discrepancy in time. The accuracy score obtained for both classifiers is 1. In the bottom part of figure \ref{fig:pred_sensors}, we can observe that the robot knows quite early which decision it will take at the next loop while we could expect that it would take its decision in the yellow corridor in figure \ref{fig:t-maze}. Here, we see that if the robot just turned right, the reservoir switches its internal state to go left next time only a few dozen time steps after. We tried the same classifiers but instead of the reservoir states as input, we used the sensors values. The results are shown in the figure \ref{fig:pred_sensors}. As expected, the classifications fail with an accuracy score of 0.57 for SVM and 0.43 for KNN; this randomness can be seen in both figures. Thus, we showed that by simply observing the internal states of the reservoir, it is possible to predict its next prediction. In essence, this is a proof of concept to show that second-order or observer networks, mimicking the role of the regions of the prefrontal cortex implementing contextual rules, can consolidate information linking sensory information to motor actions, to develop relevant contextual representations.  
\begin{figure}
    \centering
    \includegraphics[width=\linewidth]{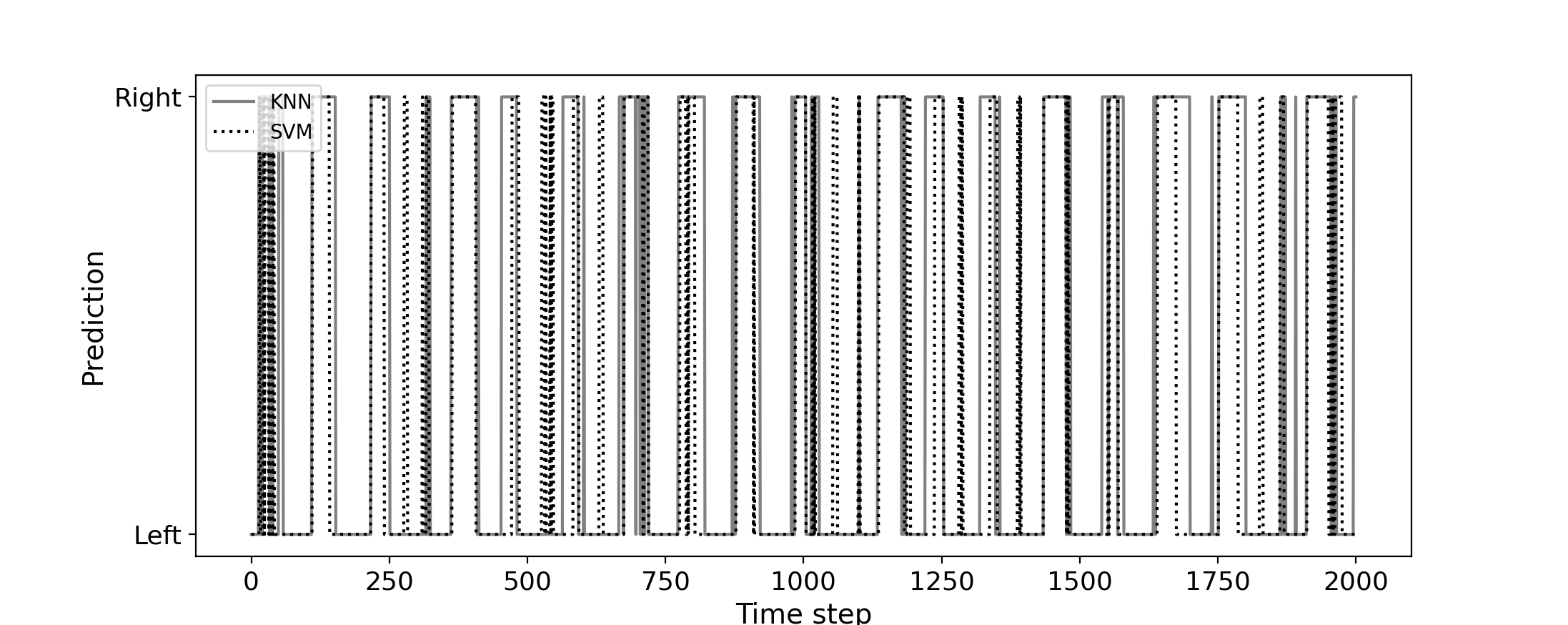}
    \includegraphics[width=\linewidth]{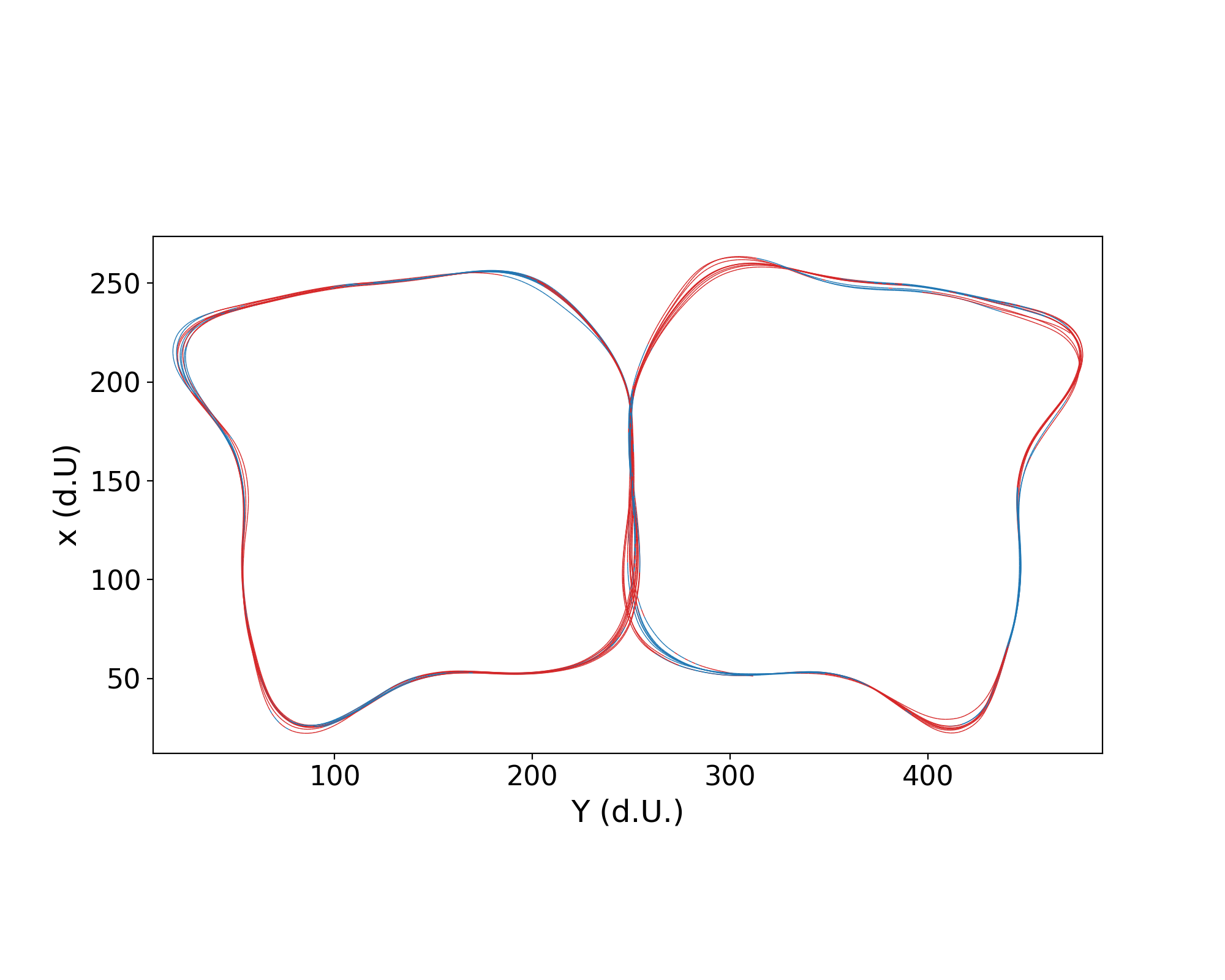}
    \caption{Prediction from sensors during 2000 time steps. Top figure shows the prediction of the KNN and SVM classifier, bottom figure shows the SVM prediction along the trajectory.}
    \label{fig:pred_sensors}
\end{figure}
\begin{figure}
    \centering
    \includegraphics[width=\linewidth]{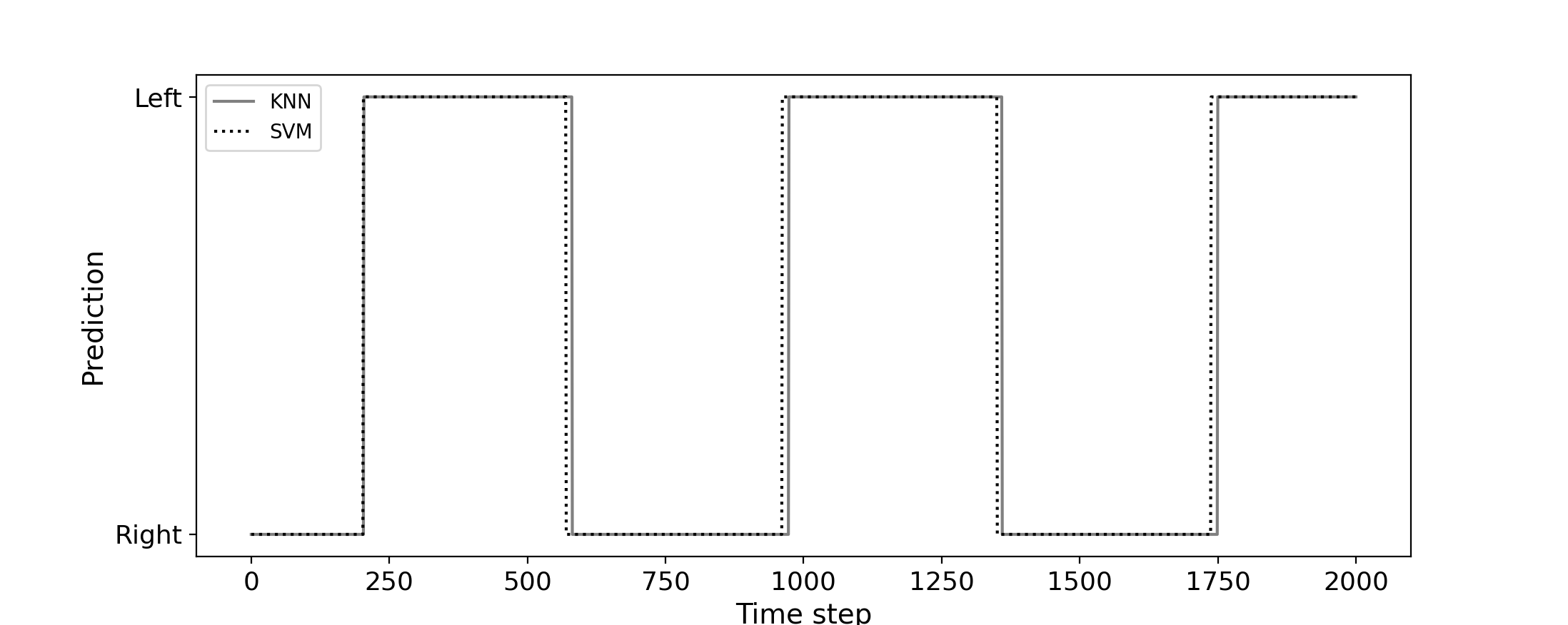}
    \includegraphics[width=\linewidth]{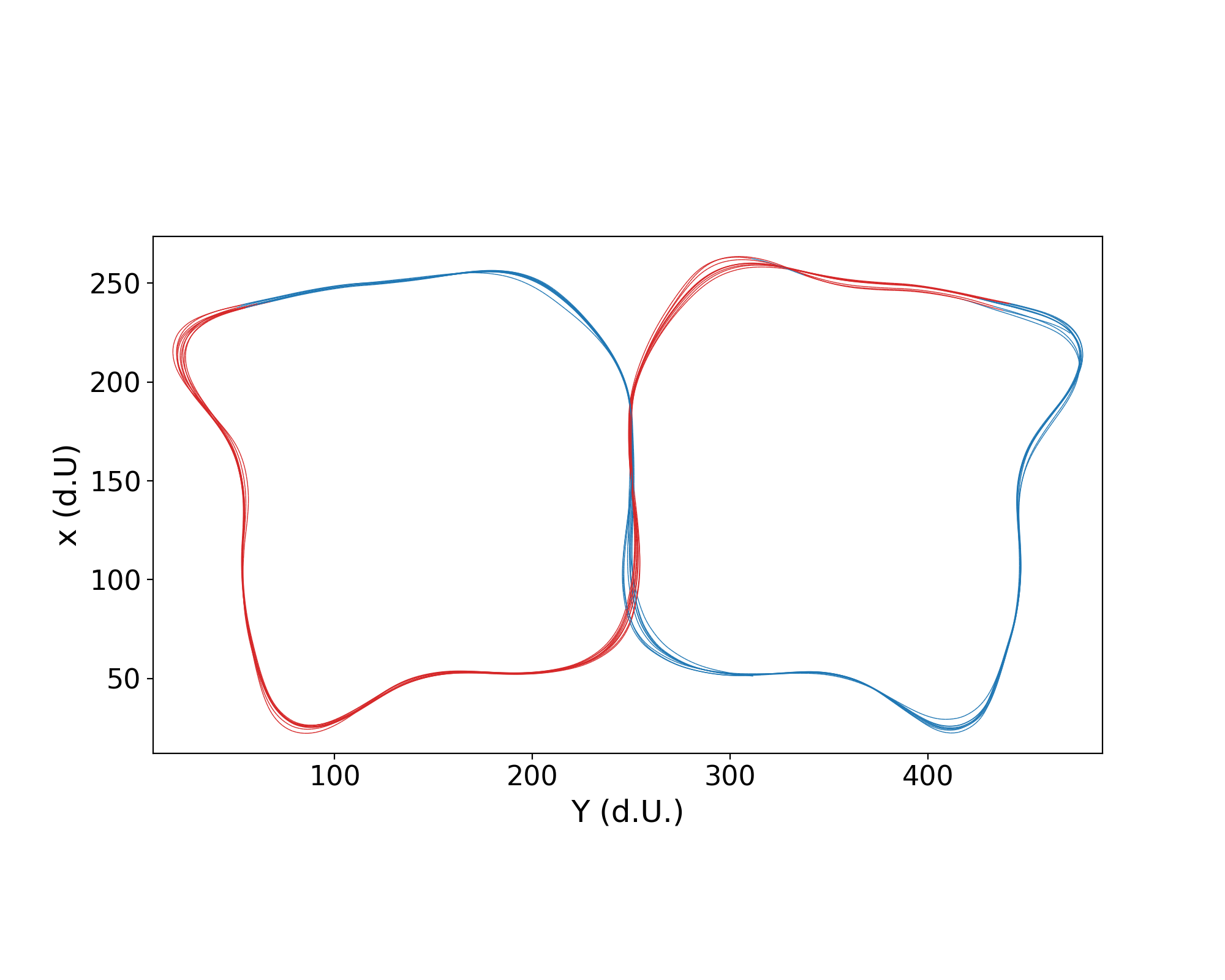}
    \caption{Prediction from reservoir state during 2000 time steps. Top figure shows the predictions of the KNN and SVM classifier. Bottom figure shows the SVM prediction along the trajectory.}
    \label{fig:pred_states}
\end{figure}
%
%\begin{figure}[h!]
%    \centering
%    \includegraphics[width=\linewidth]{Figures/SVM_plot_sv.png}
%    \caption{Prediction of the SVM classifier during 2000 time steps. The figure on top shows the predictions when the input of the SVM is the sensors values. The figure at the bottom shows the predictions when the input of the SVM is the reservoir state.}
%    \label{fig:SVM_pred_spatially}
%\end{figure}
%
%\begin{figure}[!h]
%    \centering
%    \includegraphics[width=\linewidth]{Figures/classifier_prediction_sensors.png}
%    \caption{Predictions of the classifiers from the reservoir states during 2000 time steps in the 8-shape trajectory.  The plots in dot show the SVM predictions while the plot in solid line the KNN predictions. The inputs of the classifiers are the sensors values.}
%    \label{fig:classifier_pred}
%\end{figure}
%
%\begin{figure}[!h]
%    \centering
%    \includegraphics[width=\linewidth]{Figures/classifier_prediction_reservoir_states.png}
%    \caption{Predictions of the classifiers from the reservoir states during 2000 time steps in the 8-shape trajectory.  The plots in dot show the SVM predictions while the plot in solid line the KNN predictions. The inputs of the classifiers are the reservoir states.}
%    \label{fig:classifier_pred}
%\end{figure}
%
Since the state space of the dynamic reservoir is high-dimensional, using the Principal Component Analysis (PCA) on the states, we investigated if it is possible to observe sub-space attractors. The result for the PCA analysis is presented in figure \ref{fig:PCA}, where PCA was applied for 5000 time steps, which corresponds to 7 8-loops. The figure shows two symmetric sub-attractors, which are linearly separable, that correspond to the two parts of the 8-shape trajectory. 
\begin{figure}
    \centering
\includegraphics[width=1\linewidth]{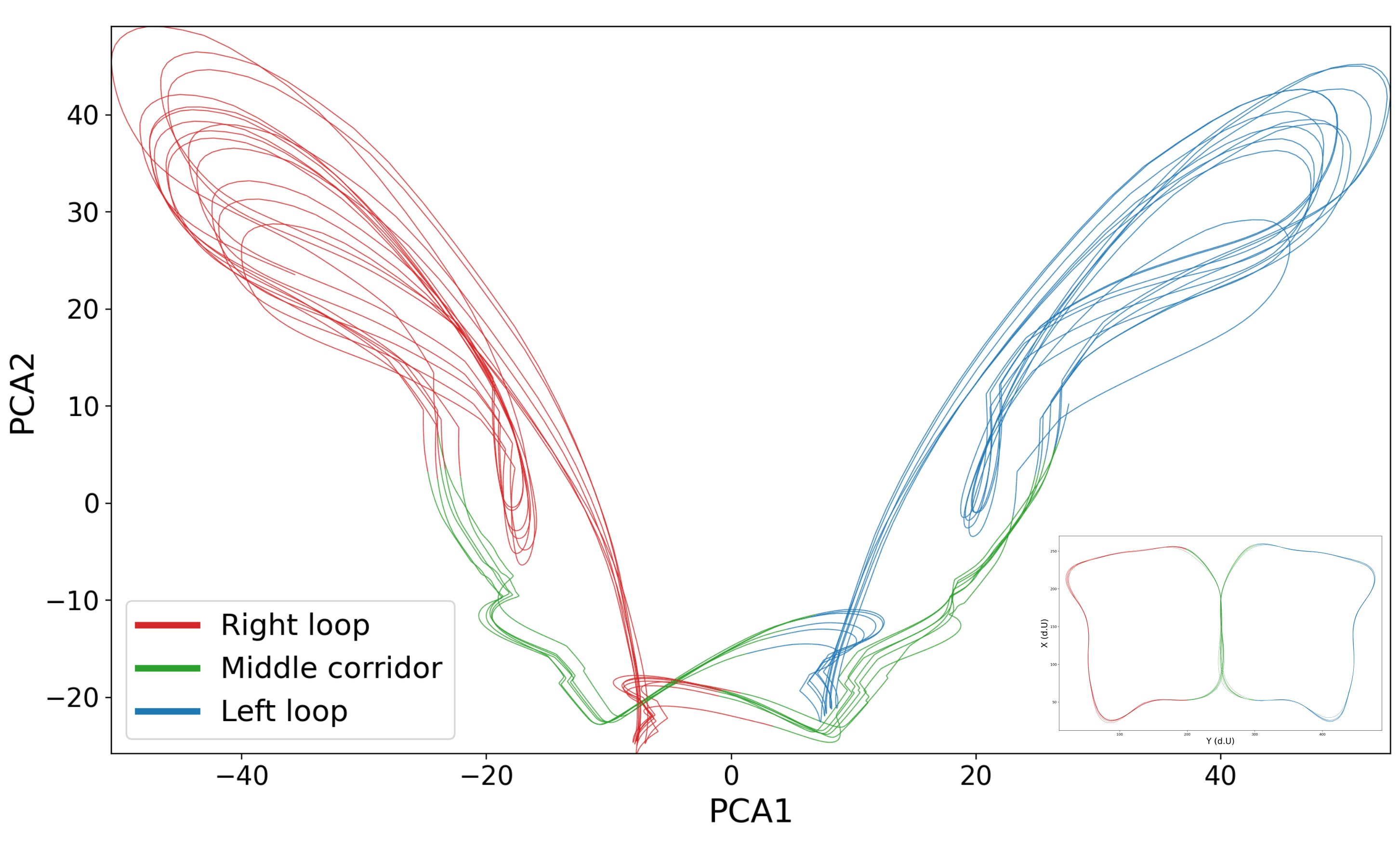}
    \caption{The first two principal components of the reservoir state space after applying PCA on the reservoir states. On the bottom right is the corresponding map of the positions of the robot in the maze.}
    \label{fig:PCA}
\end{figure}

\subsection{Explicit rules with contextual inputs}

Finally, we demonstrate that although the ESN can learn a motor sequence without contextual inputs, it is limited by its internal representation to learn more complex sequences which may require a longer memory. Adding contextual or explicit information about the rule (which we propose are representations developed by the prefrontal cortex over time) can then bias the ESN to follow any arbitrary trajectory as in \ref{fig:positions}. With the additional contextual inputs, the ESN is able to reproduce the standard 8 sequence (the performance is shown in table \ref{tab:score2}) but can also achieve more complex tasks by sending to it the proper contextual inputs. One example can be seen in figure \ref{fig:positions}: the top graph shows the positions of the bot while making the standard 8 sequence [ABABABAB...], the bottom one shows that the bot was able to achieve a more complex sequence  [AABBAABBAABB...]. 
 \begin{table}
    \centering
    \begin{tabular}{l l l l}
    \hline
    \multicolumn{4}{l}{Performance of the ESN for 50 runs}\\
    \hline
    \multicolumn{2}{l}{\(NRMSE\)} & \multicolumn{2}{l}{\(R^2\)} \\
    \hline
    Mean &  0.0050  &  Mean &   0.9997 \\
    Variance &  1.1994e-07 & Variance & 2.0220e-09 \\
    \hline
    \end{tabular}
    \caption{\(NRMSE\) and \(R^2\) score of the ESN with the two additional contextual inputs}
    \label{tab:score2}
\end{table}
\begin{figure}
    \centering
    \includegraphics[width=\linewidth]{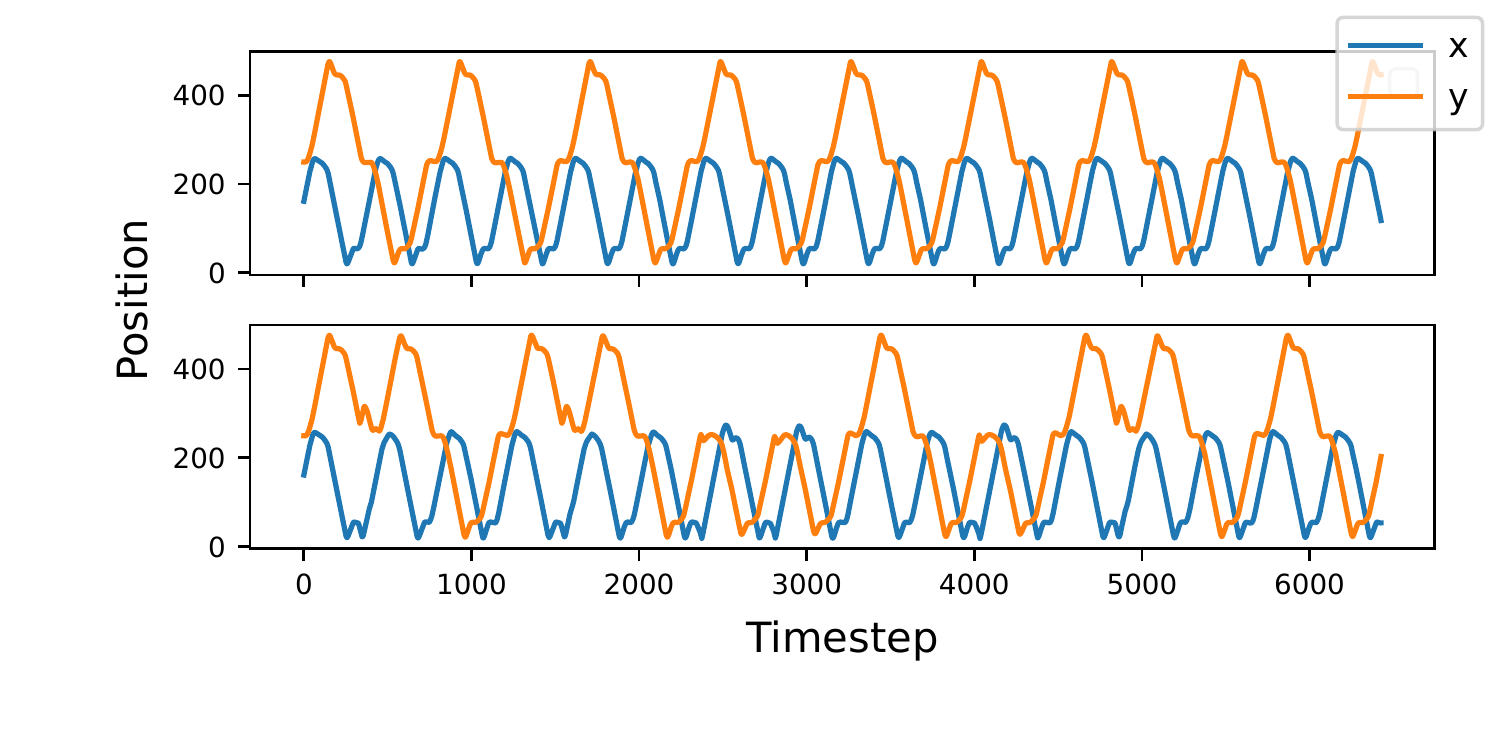}
    \caption{The coordinates of the agent for 7000 timesteps in the prediction phase. The plots in blue show the x axis coordinates while the ones in red show the y axis coordinates. The figure on top shows the results for the standard 8 sequence [ABABAB..], the figure at the bottom shows the results for a randomly generated sequence [AABAABBBABBAABAB], where 'A' is the left loop and 'B' is the right loop.}
    \label{fig:positions}
\end{figure}

\section{Discussion}

Using a simple reservoir model that learns to follow a specific path, we have shown how the resulting behavior could be interpreted as an alternating behavior by an external observer. However, we've also shown that from the point of view of the model and in the absence of associated cues, this behavior cannot be interpreted as such. Instead, the behavior results from the internal dynamics of the reservoir (and the learning procedure we implemented). Without external cues, the model is unable to escape its own behavior and is trapped inside an attractor. Only the cues can provide the model with the necessary and explicit information that in turn allows to bias its behavior in favor of option A or option B. 

From a neuroscience perspective, as developed in more details in \cite{koechlin_evolutionary_2014}, it can be proposed that the reservoir model in the first experiment implements the premotor cortex learning sensorimotor associations in the anterior cortex. In the first experiment, this is made by supervised learning in a process of learning by imitation. In a different protocol, this is also classically be done by reinforcement learning, involving another region of the anterior cortex, the anterior cingulate cortex, manipulating prediction of the outcome. Whereas both regions of the anterior cortex and present in mammals, \cite{koechlin_evolutionary_2014} reports that another region, the lateral prefrontal cortex, is unique in primates and has been developed to implement the learning of contextual rules and to possibly act in a hierarchical way in the control of the other regions. We have proposed an elementary implementation of the lateral prefrontal cortex in the second experiment, adding explicit contextual inputs as a basis to form contextual rules. It was accordingly very important to observe that it was then possible to explicitly manipulate the rules and form flexible behavior, whereas in the previous case, rules were implicitly present in the memory but not manipulable.

This simple model shows that the interpretation of the behavior by an observer and the actual behavior might greatly differ even when we can make accurate prediction about the behavior. Such prediction can be incidentally true without actually revealing the true nature of the underlying mechanisms. Based on the reservoir computing framework which can be invoked for both premotor and prefrontal regions, we have implemented models which are structurally similar (as it is the case for that regions) and we have shown that a simple difference related to their inputs can orient then toward implicit or explicit learning as respectively observed in the premotor and lateral prefrontal regions. It will be important in future work to see how these regions are associated to combine both modes of learning and switch from on to the other depending on the complexity of the task.  

% \bibliography{draft.bib}

\end{document}